\documentclass{article}
\usepackage{amssymb}
\usepackage[table]{xcolor}

\usepackage[preprint]{corl_2026} % Uncomment for pre-prints (e.g., arxiv); This is like ``final'', but will remove the CORL footnote.

\usepackage[pdftex]{graphicx}
\usepackage{amsmath}
\usepackage{amssymb}  % assumes amsmath package installed
\usepackage{subfigure}
\usepackage{subcaption}
\usepackage{multirow}
\usepackage{array,booktabs}
\usepackage{diagbox}
\usepackage{balance}
\usepackage{xspace}
\usepackage{algorithm}
\usepackage{algpseudocode}
\usepackage{adjustbox}
\usepackage{romannum}
\usepackage{cleveref}
\crefname{figure}{Fig.}{Figs.}
\Crefname{figure}{Fig.}{Figs.}
\usepackage{rpm_SIunits}
\usepackage{rpm_acronyms}
\usepackage{rpm_math}
\usepackage{rpm_misc}
\usepackage{bbm}

\usepackage{textcomp}

\usepackage{soul,color}
\usepackage{lipsum}
\usepackage{dsfont}

\usepackage{xcolor}

\definecolor{imlprcolor}{HTML}{f3f7fc} % Light blue for ImLPR row
\definecolor{bestcolor}{HTML}{def3e6} % Light green for best results
\definecolor{secondbestcolor}{HTML}{ecf8f1} % Very light green for second-best results

\usepackage{amssymb}% http://ctan.org/pkg/amssymb
\usepackage{pifont}% http://ctan.org/pkg/pifont
\newcommand{\cmark}{\ding{51}}%
\newcommand{\xmark}{\ding{55}}%
\usepackage[font=small]{caption}
\usepackage{tikz} % \checkmark
\usepackage{subcaption}
\usepackage{caption}
\title{\LARGE \bf 
RoMo-GS: Efficient Online 3D Gaussian Splatting for Robot Mapping via Exposure-Integrated Motion Blur
}

\author{
  Sangwoo Jung${}^{1,2}$, Di Deng${}^1$, Jiahui Fu${}^1$, Laura Herlant${}^{1*}$, Ayoung Kim${}^{2*}$, Yafei Hu${}^{1*}$\\
  ${}^1$RAI Institute, United States\\
  ${}^2$Department of Mechanical Engineering, Seoul National University, South Korea\\
  ${}^*$Co-corresponding Authors
}

\begin{document}
\maketitle

%===============================================================================

\begin{figure*}[h]
    \centering
    \vspace{-10mm}
    \includegraphics[width=\textwidth]{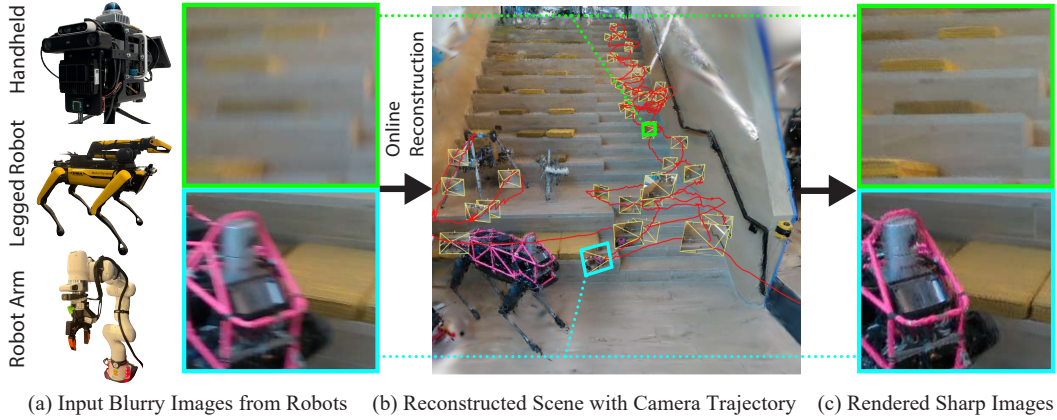}
    \vspace{-5mm}
    \caption{We introduce \textbf{RoMo-GS}, an efficient online 3D Gaussian Splatting system for real-world robotic mapping. Given a stream of motion-blurred frames captured by robots and front-end vision tracker outputs (a), RoMo-GS produces a compact 3DGS map (b) and sharp renderings online (c). }
    \label{fig:teaser}
\vspace{-5mm}
\end{figure*}

\begin{abstract}
Online 3D Gaussian Splatting (3DGS) is emerging as a practical way to build a representation for robotics applications. However, deploying it on real-world platforms exposes multiple challenges. Rapid maneuvers of a robot induce motion-blurred images, which corrupt the reconstructed geometry and cause camera pose drift. Furthermore, the number of Gaussians grows without bound during continuous online mapping. In this study, we present \textbf{RoMo-GS}, an online 3DGS system designed to address these challenges. Instead of averaging $N$ discrete sub-frames as in prior motion blur-aware 3DGS, we analytically integrate each projected Gaussian over the exposure interval, yielding an $\mathcal{O}(1)$ complexity per Gaussian. Building on this, we bound the Gaussians' growth with a Hessian-based pruning score, while jointly refining pose, velocity, and Gaussians with Adam and an online LM step over the Gaussians. On multiple public datasets and a self-collected dataset, RoMo-GS matches or exceeds offline deblurring 3DGS methods in rendering quality while running 15--150$\times$ faster, and operates online at competitive quality with about 20\% fewer Gaussians than the online 3DGS baseline. 
\end{abstract}

\keywords{Online 3D Gaussian Splatting, Motion Deblurring, Online Density Control, Robotic Mapping}

%\onecolumn
% \maketitle
% \thispagestyle{empty}
% \pagestyle{empty}

% \input{0_abstract}
\section{Introduction}
\label{sec:intro}
Online dense 3D reconstruction is essential for robotics, enabling downstream tasks such as autonomous navigation and reliable interactions. While implicit representations such as \ac{NeRF}~\cite{mildenhall2021nerf} have advanced novel-view synthesis, their slow rendering ($\sim$0.03 FPS) limits their online use on real-world robots. \ac{3DGS}~\cite{kerbl3Dgaussians} addresses this by replacing the implicit field with explicit Gaussian primitives and a differentiable rasterizer, achieving real-time rendering ($\sim$100 FPS). Moving the rasterizer from controlled benchmarks onto actual robots, however, exposes the Gaussians to less controlled conditions. Rapid maneuvers and mechanical impacts induce motion blur in the input frames, corrupting the reconstructed geometry and causing trajectory drift, while the Gaussians grow unbounded along the trajectory until memory exhaustion.

As \ac{3DGS} migrates to real-world robotic systems, contact impacts, vibration, and rapid maneuvers produce motion-blurred frames that corrupt the reconstructed geometry. Prior motion blur-aware \ac{3DGS} methods~\cite{zhao2024bad, seiskari2024gaussian, lee2025comogaussian} mitigate this numerically, approximating the exposure integral by averaging $N$ discrete subframes and incurring an $N\times$ training-cost overhead that is incompatible with online operation. We eliminate this trade-off by analytically evaluating the exposure-time integral of each projected Gaussian under a linear motion model, yielding $\mathcal{O}(1)$ complexity per Gaussian. 

Beyond rendering, an online \ac{3DGS} system on a real-world robot faces remaining challenges. As the robot continuously explores the environment, new Gaussians are spawned with each incoming frame. Without a principled mechanism to prune the redundant or uninformative Gaussians, the number of Gaussians grows without bound and exhausts the robot's memory budget. Naive pruning, on the other hand, degrades overall reconstruction quality. Furthermore, during rapid maneuvers, pose tracking may become unstable, introducing pose drift and corrupting structural integrity. 

To address these constraints, we propose \textbf{RoMo-GS}, an online \ac{3DGS} pipeline designed for real-world robot deployment. To handle motion blur without multi-sampling, we analytically evaluate intra-exposure camera motion. To bound memory, we introduce an online Gaussian pruning score based on the opacity-axis diagonal of the Gauss-Newton Hessian. Finally, to resolve pose drift by tightening the tracking and mapping, we run a joint Adam loop over camera states and Gaussian parameters, followed by an online \ac{LM} step on the Gaussians. We evaluate RoMo-GS on multiple public datasets containing real-world and synthetic blurry scenes, as well as a custom multi-platform dataset. In summary, our contributions are as follows.
\begin{itemize}
    \item \textbf{Exposure-Integrated Motion Blur.} We evaluate the exposure-time integral of each projected Gaussian's motion in terms of the \ac{ERF}, replacing the discrete numerical approximations of prior approaches with our analytically evaluated formulation. 

    \item \textbf{Single-Sample Blur Rasterization.} We integrate the proposed motion-blur model into the CUDA rasterizer to render motion-blurred frames in a single pass with $\mathcal{O}(1)$ complexity per Gaussian, removing the $N\!\times$ rasterization overhead of multi-sample averaging approaches. 

    \item \textbf{Online Optimization and Density Control.} We jointly optimize pose, velocity, and Gaussians with Adam, followed by an online \ac{LM} step on the Gaussians, and bound the map size with a Hessian-based pruning score that retains $\sim$20\% fewer Gaussians at comparable quality.
\end{itemize}

\section{Related Work}
\label{sec:relatedwork}

\subsection{Online Differentiable Rendering for Robotics}
Driven by modern GPUs, neural rendering has emerged as essential for 3D reconstruction. NeRF~\cite{mildenhall2021nerf} produces high-quality novel views via ray-tracing through an implicit \ac{MLP}, and subsequent works accelerate its training and rendering~\cite{chen2022tensorf, fridovich2022plenoxels, garbin2021fastnerf, yu2021plenoctrees}. \ac{3DGS}~\cite{kerbl3Dgaussians} achieves real-time rendering through an explicit representation and differentiable Gaussian rasterization.

Leveraging the speed of \ac{3DGS}, recent work has extended to online reconstruction. With RGB-D input, MonoGS~\cite{matsuki2024gaussian} and SplaTAM~\cite{keetha2024splatam} couple \ac{3DGS} with pose tracking; RTG-SLAM~\cite{peng2024rtg} accelerates optimization through a compact representation. GS-ICP SLAM~\cite{ha2024rgbd} fuses \ac{GICP} to enhance tracking speed and mapping quality, and Photo-SLAM~\cite{huang2024photo} adopts Gaussian-Pyramid learning to accelerate optimization. With RGB input, HI-SLAM2~\cite{zhang2025hi} integrates learning-based dense \ac{SLAM} for improved geometry, and MODP~\cite{wu2025monocular} presents hierarchical Gaussian management and global consistency optimization.

Still, deploying these systems on real robots breaks the assumption of high-quality inputs: vibration, contact impacts, and rapid maneuvers produce motion-blurred frames that degrade the reconstruction and drive pose drift, while unbounded Gaussian growth during online mapping inflates memory. 

\subsection{Deblurring Neural Rendering}
Prior works have explored deblurring strategies within the neural rendering framework to mitigate motion blur. Early approaches, such as Deblur-NeRF~\cite{ma2022deblur}, adopt 2D blind-deblurring with a learned blur kernel. Deblurring 3DGS~\cite{lee2024deblurring} and BAGS~\cite{peng2024bags} extend the kernel-based formulation with \ac{MLP}- and \ac{CNN}-based kernels. Unblur-SLAM~\cite{zhang2026unblur} combines a feed-forward deblurring network with multi-view kernel learning. Still, kernel-based approaches treat blur as an image-space artifact and lack a physical interpretation of the camera's kinematics.

To preserve geometric consistency, methods such as DP-NeRF~\cite{lee2023dp} model intra-exposure camera motion as a rigid-body trajectory. Building on this, BAD-NeRF~\cite{wang2023bad} and BAD-Gaussians~\cite{zhao2024bad} model camera trajectories by interpolating with spline-based methods. \ac{GSOM}~\cite{seiskari2024gaussian} derives a pixel-velocity motion blur model under a screen-space approximation, while CoMoGaussian~\cite{lee2025comogaussian} extends kinematic deblurring to continuous motion by leveraging neural \ac{ODEs}. These kinematic motion-based approaches simulate blur by rendering and averaging $N$ sub-frames along the trajectory and scale the training cost as $\mathcal{O}(N)$, prohibitive for online robot deployment. Instead, we leverage the explicit representation of \ac{3DGS} to analytically compute motion blur in a single rasterization pass. 

\subsection{Recent Advances in Enhancing 3DGS}
To achieve fast convergence and compact scene representation in \ac{3DGS}, recent work has refined optimization and pruning. 3DGS-LM~\cite{hollein20253dgs} appends second-order \ac{LM} steps after the Adam loop to accelerate convergence, while Speedy-Splat~\cite{hanson2025speedy} and FastGS~\cite{ren2025fastgs} compact the Gaussians via offline score-based pruning and tighten per-Gaussian pixel coverage. Although effective offline, they rely on access to the converged scene or a full backward pass over all training views, so directly applying them to online \ac{3DGS} introduces additional compute overhead and degrades render quality. 

A complementary line of work~\cite{Lin2026DA3, Wang2026TTTLRM} explored feed-forward Gaussian reconstruction, leveraging recent 3D vision foundation models~\cite{Lin2026DA3, Wang2025VGGT}. They exploit robust depth priors to suppress the floater artifacts that plague optimization-based pipelines. Still, their fixed Gaussian count removes the adaptive density control, which limits capturing high-frequency textures. We adopt the score-based pruning and the depth-prior bootstrapping, reformulating both to operate within an online system.

% Beyond optimization-based mapping, several works explored feed-forward Gaussian reconstruction, driven by recent advancements in 3D vision foundation models~\cite{Wang2025VGGT, Lin2026DA3}. These methods~\cite{Lin2026DA3, Wang2026TTTLRM} leverage robust depth priors to mitigate the geometric ambiguity that produces floater artifacts in traditional optimization-based approaches. However, their fixed primitive count lacks the adaptive density control of traditional methods, which limits their ability to capture high-frequency textures. 

% \lhnote{Given the structure of the contributions, should we also include a section in related work on joint camera pose, velocity, and GS optimization?}

% several attempts have modeled precise blur kernels across baselines such as NeRF~\cite{mildenhall2021nerf} and 3DGS~\cite{kerbl3Dgaussians}. Deblurring 3DGS~\cite{lee2024deblurring} adjusts Gaussian covariance and mean using a small \ac{MLP} layer to generate blurry images during training, while BAGS~\cite{peng2024bags} proposes a \ac{CNN}-based multi-scale degradation kernel equipped with blur masking. 

\section{Methods}

\begin{figure*}[t!]
    \centering
    \includegraphics[width=\textwidth]{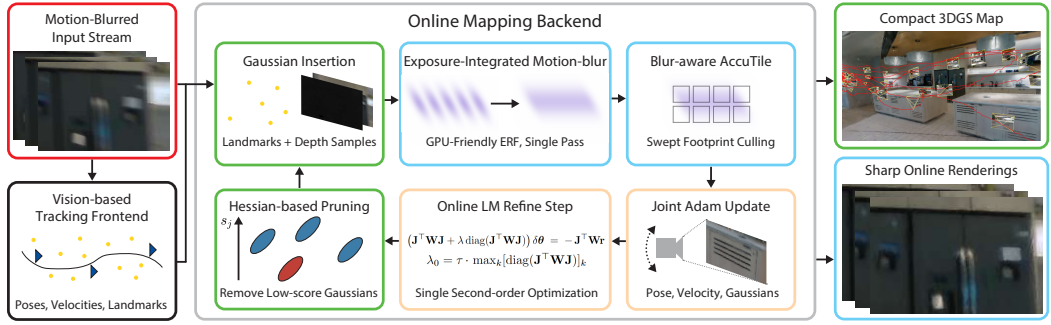}
    \caption{Overview of \textbf{RoMo-GS}. Given inputs from the vision-based tracker, we spawn initial Gaussians with the landmarks and depth samples. The exposure-integrated motion blur model renders a blurry image in a single rasterization. A joint Adam loop and an \ac{LM} step refine the camera state and Gaussians, while Hessian-based score prunes redundant Gaussians. The pipeline produces a compact 3DGS map and sharp online renderings. }
    \label{fig:overview}
\vspace{-6mm}
\end{figure*}

Our system is tracker-agnostic and compatible with any visual \ac{SLAM}, including OKVIS2-X~\cite{Boche2025okvis2x}. We adopt ORB-SLAM3~\cite{campos2021orb} as the front-end tracker and build on MODP~\cite{wu2025monocular}, an online monocular \ac{3DGS} system. For optimization, standard \ac{3DGS} loss $\mathcal L=(1-\lambda_\text{ssim})\mathcal L_1(\hat I,I)+\lambda_\text{ssim}(1-\mathrm{SSIM}(\hat I,I))$ with $\lambda_\text{ssim}=0.2$ is used. An overview of our method is shown in \figref{fig:overview}. 

\subsection{Analytic Motion Blur Modeling for 3DGS}
\label{subsec:blur}

\paragraph{Image under Camera Motion.}
The observed image of frame $i$ captured by a moving camera integrates the rasterized color $\mathbf{C}_i(\mathbf{x}, \mathbf{T}(t_i))$ evaluated at pixel $\mathbf{x}$ under pose $\mathbf{T}(t)\in\mathrm{SE}(3)$:
\begin{equation}
\small
    \bar{\mathbf{C}}_i(\mathbf{x})
    =
    \frac{1}{t_e} \int_{-t_e/2}^{t_e/2}
    \mathbf{C}_i(\mathbf{x}, \mathbf{T}(t_i + \tau))~ d\tau,
    \label{eq:blur_integral}
\end{equation}
with exposure $t_e$. Assuming constant linear and angular velocities $(\mathbf{v}_i,~\boldsymbol{\omega}_i)$ during the exposure, $\mathbf{T}(t_i+\tau)$ admits the standard $\mathrm{SE}(3)$ linearization 
$[\mathbf{R}_i\exp(\tau[\boldsymbol{\omega}_i]_\times) ~ | ~ \mathbf{p}_i + \mathbf{R}_i \mathbf{v}_i \tau]$ around $( \mathbf{R}_i, \mathbf{p}_i )$. Still, direct rasterization of Eq.~\eqref{eq:blur_integral} is infeasible as the integral has no closed-form solution. 

\paragraph{Screen-Space Approximation.}
Standard \ac{3DGS} represents the scene with 3D Gaussians $\{\boldsymbol{\mu}_j, \boldsymbol{\Sigma}_j, o_j, \mathbf{c}_j\}$, projected to screen-space $(\boldsymbol{\mu}'_{i,j},~\boldsymbol{\Sigma}'_{i,j})$~\cite{zwicker2002ewa}. \ac{GSOM}~\cite{seiskari2024gaussian} restricts the camera motion to the projected means so that each $\boldsymbol\mu'_{i,j}$ moves linearly during the exposure with pixel velocity $ \mathbf{v}'_{i,j} = -J_i (\boldsymbol{\omega}_i \times \hat{\boldsymbol{\mu}}_{i,j} + \mathbf{v}_i)$, where $\hat{\boldsymbol{\mu}}_{i,j} = \mathbf{R}^\top_{i} (\boldsymbol{\mu}_j - \mathbf{p}_i)$ and $J_i$ is the affine projection Jacobian at $\hat{\boldsymbol{\mu}}_{i,j}$. Under this, \ac{GSOM} averages the standard \ac{3DGS} $\alpha_{i,j} (\mathbf{x}, \boldsymbol{\mu}')=o_j \exp(-\frac{1}{2}(\boldsymbol{\mu}'-\mathbf{x})^\top(\boldsymbol{\Sigma}'_{i,j})^{-1} (\boldsymbol{\mu}'-\mathbf{x}))$ over $N_\text{blur}$ uniform trajectory samples $\tau_k=((k-1)/(N_\text{blur}-1)-1/2)t_e$:
\begin{equation}
\small
    \bar{\mathbf{C}}_i(\mathbf{x})
    =
    \frac{1}{N_\text{blur}}\sum^{N_\text{blur}}_{k=1}\sum_j \mathcal{T}_j^{(k)} \mathbf{c}_{i,j} \alpha_{i,j}(\mathbf{x}, \tilde{\boldsymbol{\mu}}'_{i,j} (t_i + \tau_k)), 
    \qquad
    \tilde{\boldsymbol{\mu}}'_{i,j} (t_i + \tau) = \boldsymbol{\mu}'_{i,j} (t_i) + \mathbf{v}'_{i,j} \tau, 
    \label{eq:multi_sample}
\end{equation}
where $\mathcal{T}_j^{(k)}=\prod_{l<j}(1 - \alpha_{i,l}(\mathbf{x}, \tilde{\boldsymbol{\mu}}'_{i,l} (t_i + \tau_k)))$ is the front-to-back transmittance at sample $k$. Still, the rasterization scales as $\mathcal{O}(N_\text{blur})$~\cite{seiskari2024gaussian}, resulting in a proportional increase in running time. 

\begin{figure*}[t!]
    \centering
    \includegraphics[width=0.8\textwidth]{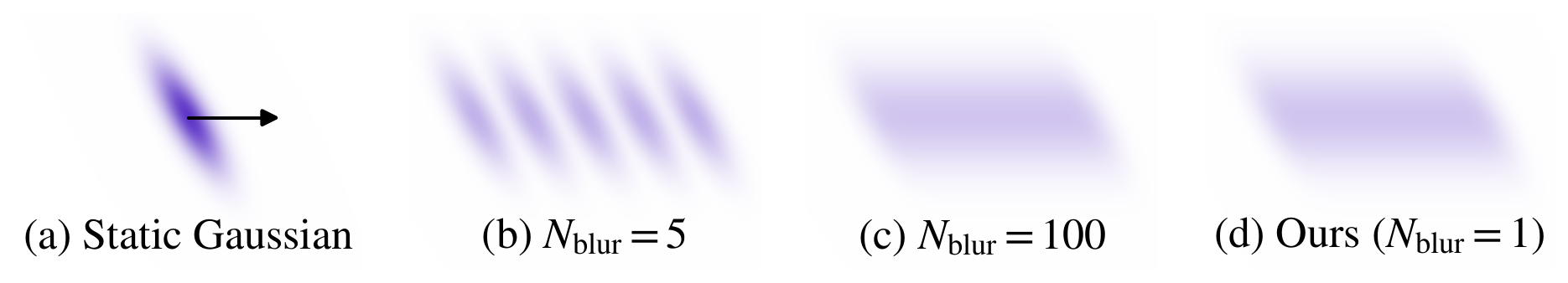}
    \vspace{-2mm}
    \caption{Comparison with multi-sample. (a) Static Gaussian with a motion arrow. (b) $N_\text{blur}=5$, visible discretization. (c) $N_\text{blur}=100$, smoother, but expensive. (d) Our model results same as (c) at a single sample. }
    \label{fig:erf_blur}
\vspace{-5mm}
\end{figure*}

\paragraph{Exposure-Integrated Motion Blur.}
We retain the screen-space approximation but analytically evaluate the exposure-time integral using the GPU-friendly \ac{ERF}, replacing the discrete $N_\text{blur}$-sample average. Letting $\boldsymbol{\delta} = \boldsymbol{\mu}'_{i,j} - \mathbf{x}$, the time-averaged opacity contribution of Gaussian $j$ at pixel $\mathbf{x}$ is
\begin{equation}
\small
    \bar{\alpha}_{i,j}(\mathbf{x})
    =
    \frac{o_j}{t_e}\int^{t_e/2}_{-t_e/2}\exp \Bigl(-\frac{1}{2}(\boldsymbol{\delta}+\mathbf{v}'_{i,j}\tau)^\top(\boldsymbol{\Sigma}'_{i,j})^{-1}(\boldsymbol{\delta}+\mathbf{v}'_{i,j}\tau) \Bigr)d\tau.
    \label{eq:erf_integral}
\end{equation}
With $ a = {\mathbf{v}'_{i,j}}^\top {\boldsymbol{\Sigma}'_{i,j}}^{-1} \mathbf{v}'_{i,j}$, $b = {\mathbf{v}'_{i,j}}^\top {\boldsymbol{\Sigma}'_{i,j}}^{-1} \boldsymbol{\delta}$, $ c = \boldsymbol{\delta}^\top {\boldsymbol{\Sigma}'_{i,j}}^{-1} \boldsymbol{\delta}$, $\tau^*=-b/a$, and $u=\sqrt{a/2}(\tau-\tau^*)$, Eq.~\eqref{eq:erf_integral} reduces to an \ac{ERF} difference-based exposure-integrated representation as
\begin{equation}
\small
    \bar\alpha_{i,j}(\mathbf x)
    =\frac{o_j}{t_e}\sqrt{\frac{\pi}{2a}}~
    \exp\bigl(-\frac{c-b^2/a}{2}\bigr)~
    [\operatorname{erf}(u_+)-\operatorname{erf}(u_-)],
    \quad u_\pm=\sqrt{\frac{a}{2}}\bigl(\!\pm\!\frac{t_e}{2}-\tau^*\bigr),
    \label{eq:erf_alpha}
\end{equation}
with the detailed derivation can be found in App.~\secref{app:blur}. The blurred image is then composited front-to-back with $\bar\alpha_{i,j}$ replacing the static $\alpha_{i,j}$, collapsing Eq.~\eqref{eq:multi_sample} into a single rasterization pass. Although the \ac{ERF} is a non-elementary integral, it is available as a hardware intrinsic on modern GPUs and evaluates with $\mathcal{O}(1)$ complexity per Gaussian. \figref{fig:erf_blur} compares the proposed motion-blur model with the multi-sample approach, while numerical stability around zero velocity is discussed in App.~\secref{app:stability}. 

\subsection{Optimization}
\label{subsec:opt}

\paragraph{Pose and Velocity Optimization.}
\label{subsubsec:pose_vel}
We adopt the pose correction of MODP~\cite{wu2025monocular} and the velocity correction of 
GSOM~\cite{seiskari2024gaussian}: active keyframe $i$ receives $\hat{\mathbf T}_i=[\operatorname{Exp}(\delta\mathbf q_i)\mathbf R_i\,|\,\mathbf p_i+\delta\mathbf p_i],~ \hat{\mathbf v}_i=\mathbf v_i+\Delta\mathbf v_i,~ \hat{\boldsymbol\omega}_i=\boldsymbol\omega_i+\Delta\boldsymbol\omega_i$ where $(\delta\mathbf p_i,\delta\mathbf q_i,\Delta\mathbf v_i,\Delta\boldsymbol\omega_i)$ are trained jointly with the Gaussians by Adam.

\paragraph{LM Refinement with Online 3DGS.}
\label{subsubsec:lm_opt}
After the Adam loop, we apply an \ac{LM} step on the Gaussian parameters with the camera frozen, following 3DGS-LM~\cite{hollein20253dgs}. We solve a Huber-weighted, diagonally damped Gauss-Newton normal equation by preconditioned conjugate gradient (App.~\secref{app:lm}). 
Since \ac{LM} is invoked on a freshly inserted keyframe rather than a converged scene, $\lambda_0$ is set following Nielsen's data-driven approach~\cite{nielsen1999damping}, proportional to the dominant entry of $\mathrm{diag}(\mathbf J^\top\mathbf W\mathbf J)$. 

\paragraph{Blur-Aware AccuTile Rasterization.}
\label{subsubsec:accutile}
We replace the AABB tile assignment of standard \ac{3DGS} with the AccuTile from Speedy-Splat~\cite{hanson2025speedy}, sweeping the screen-space ellipse row-by-row to recover exact tile ranges. To bound the time-extended representation of Eq.~\eqref{eq:erf_alpha}, we union the per-row range over three samples $\tau\in\{-t_e/2,0,t_e/2\}$ along the $\mathbf v'_{i,j}$, covering the swept hull of the Gaussian.

\subsection{Pruning and Densification}
\label{subsec:prune_densify}
MODP~\cite{wu2025monocular} prunes Gaussians by opacity and densifies via learned monocular depth~\cite{dexheimer2023learning}; we retain this design and add a Hessian-based pruning score and a depth-based Gaussian initialization. 

\paragraph{Hessian-based Online Gaussian Pruning.}
\label{subsubsec:hessian_pruning}
Inspired by the Hessian-based importance score of Speedy-Splat~\cite{hanson2025speedy}, we score each Gaussian by the per-pixel squared sensitivity of the photometric loss $L$ to its opacity $o_j$ as $s_j=\sum_i\sum_{\mathbf x}(\partial L_i/\partial o_j)^2,$ which is the opacity-axis Gauss-Newton diagonal of the training loss and is acquired from the gsplat~\cite{ye2025gsplat} backward at no extra cost (App.~\secref{app:pruning}). 
Gaussians whose loss-weighted effect covers few pixels score low and are pruned, while those observed in more than 10 keyframes are eligible.

\paragraph{Depth-based Additional Gaussian Initialization.}
\label{subsubsec:depth_initial}
Beyond the ORB-SLAM3 landmarks, we sample extra points from the depth image at pixels in high-frequency regions that are poorly explained by the current render, with probability $P(\mathbf x)=\operatorname{clip}\bigl(\|(\nabla^2 G_\sigma)*I_i\|,0,1\bigr)\cdot\operatorname{SSIM}(\hat I_i, I_i)(\mathbf x)$. The first factor targets high-frequency regions, while the second down-weights well-optimized pixels.
\section{Experiment}
\label{sec:experiment}
\subsection{Experimental Setup}
For evaluation, we use three public datasets: the Deblur-NeRF~\cite{ma2022deblur} synthetic dataset, ETH3D~\cite{schops2019bad}, and TUM RGB-D~\cite{sturm12iros}. To validate our framework with diverse real-world motion blur induced by different robots, we captured custom indoor sequences using a Boston Dynamics Spot quadrupedal robot, a Franka Emika robot arm, and a handheld platform. We report \ac{PSNR}, \ac{SSIM}, and \ac{LPIPS} scores, running speed, and the number of Gaussians. All experiments are conducted on a single NVIDIA RTX A6000 GPU. 

\subsection{Deblur 3DGS Comparison}
\begin{figure*}[!t]
    \centering
    \includegraphics[width=\textwidth]{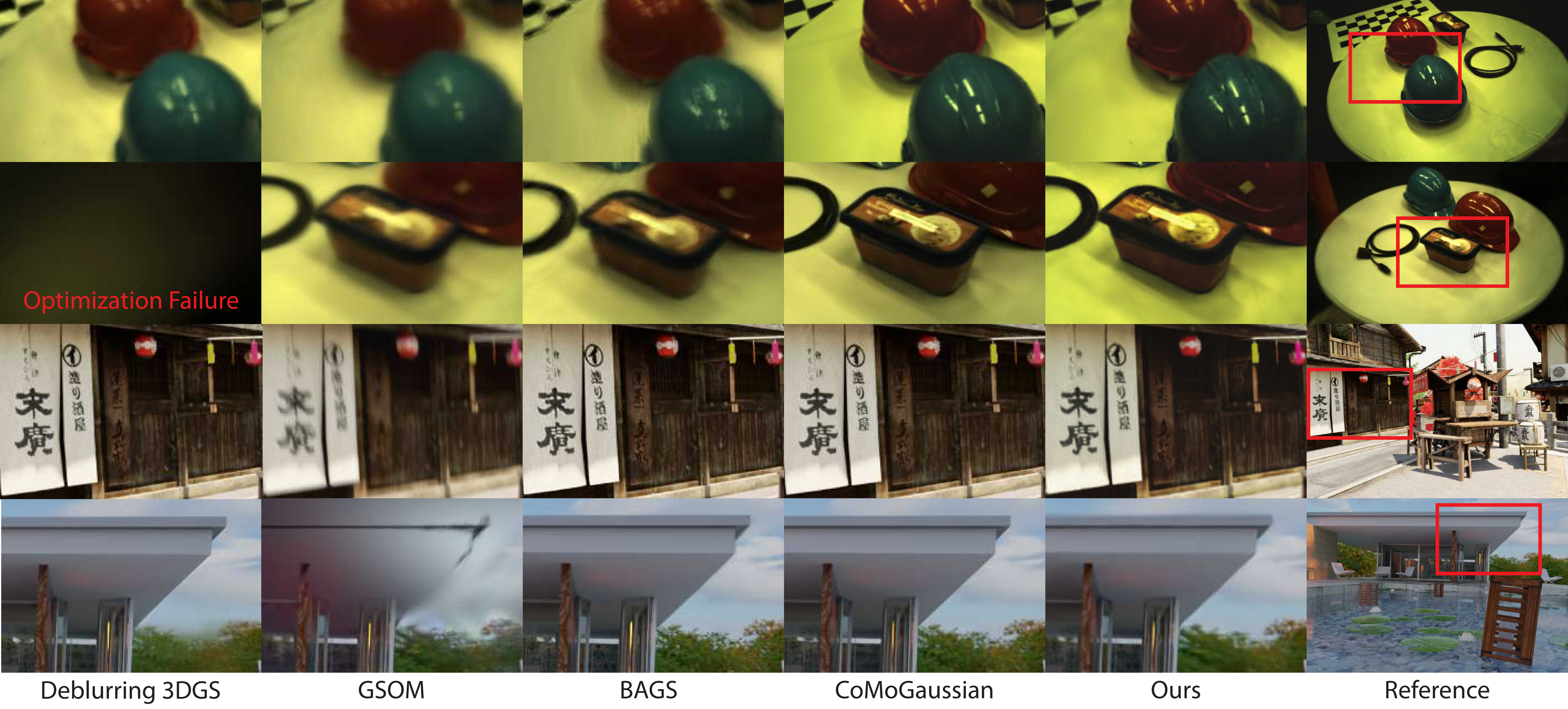}
    \caption{Top: ETH3D real-world. Bottom: Deblur-NeRF synthetic. GSOM fails to handle motion blur on both. Deblurring 3DGS and BAGS handle the synthetic case but collapse on ETH3D. CoMoGaussian is competitive on both, while Ours matches CoMoGaussian's visual quality and trains 70--150$\times$ faster online. }
    \label{fig:deblur_experiment}
\vspace{-6mm}
\end{figure*}

We compare our method against existing offline global-optimization motion-blur-aware \ac{3DGS} baselines: BAD-Gaussians~\cite{zhao2024bad}, Deblurring 3DGS~\cite{lee2024deblurring}, \ac{GSOM}~\cite{seiskari2024gaussian}, BAGS~\cite{peng2024bags}, and CoMoGaussian~\cite{lee2025comogaussian}. The evaluation uses the Deblur-NeRF~\cite{ma2022deblur} synthetic dataset and the camera-shaking sequences of ETH3D~\cite{schops2019bad}. We omit the Deblur-NeRF real-world sequences as ORB-SLAM3 fails on them. Hessian-based pruning is disabled here so that the comparison focuses on the deblurring quality. Quantitative results appear in \tabref{tab:Deblur_NeRF} and \tabref{tab:eth3d}; qualitative comparisons are shown in \figref{fig:deblur_experiment}.

On the Deblur-NeRF synthetic dataset, BAD-Gaussians attains the best overall metrics. It is the only method that performs photometric bundle adjustment, allowing it to correct the noisy ORB-SLAM3 trajectory through global optimization. We therefore treat it as an indirect upper-bound reference and primarily compare against the remaining methods, which share the same front-end poses. Among those, our framework reaches the highest \ac{PSNR} and \ac{SSIM}, demonstrating robust deblurring even under severe motion blur and inaccurate initial poses. Our method also completes running in under one minute, against 10--60 min for the offline baselines. \ac{LPIPS} is slightly higher, reflecting a marginal loss of perceptual sharpness which is expected from online, sequential optimization compared with offline, global refinement. 
% As shown in \figref{fig:deblur_experiment}, Ours matches the offline methods qualitatively.

For ETH3D, clean ground-truth is unavailable. We therefore use the outputs of BAD-Gaussians ($\sim$4 hours of offline optimization) as a reference, while noting that this introduces a bias toward multi-sample rendering. Due to the aggressive camera motion, Deblurring 3DGS, GSOM, and BAGS fail to handle motion blur in \figref{fig:deblur_experiment}. Deblurring 3DGS even fails to optimize on the second sequence, presenting a limitation of the \ac{MLP} kernel. Ours attains the best \ac{SSIM} and \ac{LPIPS} among the baselines and a \ac{PSNR} comparable to \ac{GSOM}, with the fastest optimization. Achieving the best \ac{LPIPS} on these challenging real-world sequences indicates that our motion-blur model avoids the discretization artifacts of multi-sample and remains robust under real-world motion. In \figref{fig:deblur_experiment}, Ours produces fidelity comparable to CoMoGaussian, whereas the remaining baselines yield blurry renderings.

\begin{table}[!t]
\centering
\caption{Deblur-NeRF Synthetic Dataset Experiment Results}
\label{tab:Deblur_NeRF}
\resizebox{0.8\columnwidth}{!}{
\begin{tabular}{c|c||c|c|c|c|c}
\toprule
\hline
& \textbf{BAD-Gaussians} & \textbf{Deblurring 3DGS} & \textbf{GSOM} & \textbf{BAGS} & \textbf{CoMoGaussian} & \textbf{Ours} \\ \hline
\textbf{PSNR} $\boldsymbol\uparrow$ & 28.367 & \underline{24.174} & 22.895 & 24.606 & 23.650 & \textbf{25.148} \\ \hline
\textbf{SSIM} $\boldsymbol\uparrow$ & 0.903 & \underline{0.768} & 0.775 & 0.767 & 0.748 & \textbf{0.799} \\ \hline
\textbf{LPIPS} $\boldsymbol\downarrow$ & 0.055 & \underline{0.097} & 0.274 & \textbf{0.090} & 0.102 & 0.179 \\ \hline
\textbf{Train Time (sec)}& 2073.03 & \underline{623.92} & 1431.63 & 1938.75 & 6203.25 & \textbf{41.68} \\ \hline
\textbf{\# of Gaussians}& 261,602 & 168,837 & 121,553 & 141,271 & 55,845 & 75,813 \\ 
\hline
\bottomrule
\end{tabular}
}
\vspace{-6mm}
\end{table}

\begin{table}[!t]
\centering
\caption{ETH3D Real World Dataset Experiment Results}
\label{tab:eth3d}
\resizebox{0.65\columnwidth}{!}{
\begin{tabular}{c|c|c|c|c|c}
\toprule
\hline
& \textbf{Deblurring 3DGS} & \textbf{GSOM} & \textbf{BAGS} & \textbf{CoMoGaussian} & \textbf{Ours} \\ \hline
\textbf{PSNR} $\boldsymbol\uparrow$ & 14.278 & \textbf{25.879} & 19.414 & 19.764 & \underline{25.367} \\ \hline
\textbf{SSIM} $\boldsymbol\uparrow$ & 0.545 & \underline{0.881} & 0.842 & 0.842 & \textbf{0.905} \\ \hline
\textbf{LPIPS} $\boldsymbol\downarrow$ & 0.430 & 0.186 & 0.202 & \underline{0.147} & \textbf{0.113} \\ \hline
\textbf{Train Time (sec)} & \underline{685.73} & 1648.60 & 2922.50 & 7518.50 & \textbf{117.74} \\ \hline
\textbf{\# of Gaussians} & 141,926 & 112,874 & 153,041 & 54,777 & 117,669 \\ 
\hline
\bottomrule
\end{tabular}
}
\vspace{-3mm}
\end{table}

\subsection{Online 3DGS Comparison}
\begin{figure*}[!t]
    \centering
    \includegraphics[width=\textwidth]{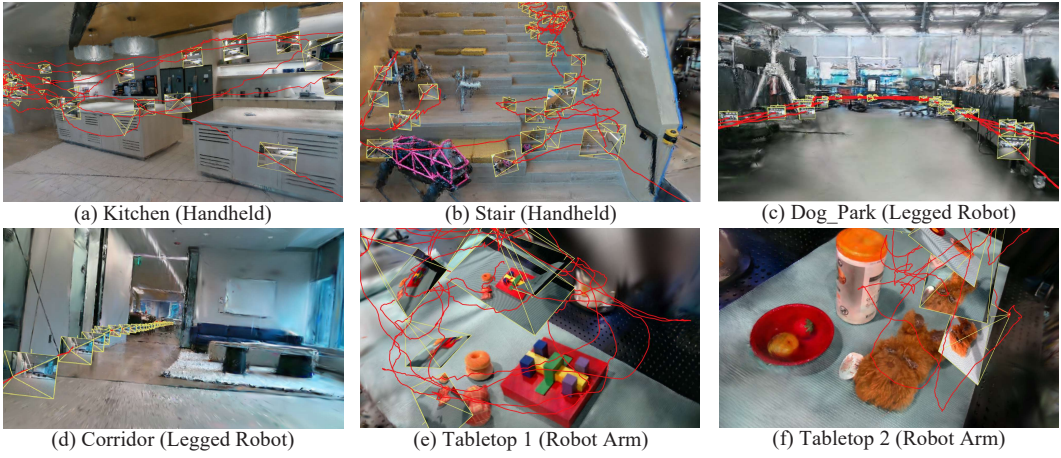}
    \caption{Scene reconstruction result of RoMo-GS on custom dataset. The red line shows the robot trajectory, including jittering due to the system's physical interaction with the environment, and the yellow squares are the camera frustums. }
    \label{fig:custom_qual}
\vspace{-5mm}
\end{figure*}

We compare our method with existing online \ac{3DGS} including RTG-SLAM~\cite{peng2024rtg}, GS-ICP~\cite{ha2024rgbd}, MonoGS~\cite{matsuki2024gaussian}, Photo-SLAM~\cite{huang2024photo}, HI-SLAM2~\cite{zhang2025hi}, and MODP~\cite{wu2025monocular}. Additionally, we include tttLRM~\cite{Wang2026TTTLRM}, a recent feed-forward Gaussian reconstruction method, for a comprehensive comparison. We evaluate on the TUM~\cite{sturm12iros} real-world dataset and our custom dataset. The deblur model is disabled in this experiment, as clean ground-truth is unavailable, and the baselines do not target motion-blurred input. Therefore, we leveraged the original input RGB images as the ground-truth reference, as in previous online Gaussian Splatting methods. Quantitative results are reported in \tabref{tab:tum} and \tabref{tab:custom}, and qualitative results on the custom dataset are shown in \figref{fig:custom_qual}. Details about the custom dataset are provided in App.~\secref{app:dataset}.

On both TUM and our custom datasets, the rendering quality of GS-ICP and Photo-SLAM remains degraded, a trade-off for their fast optimization speed. tttLRM and GS-ICP show a higher number of Gaussians than the others, indicating limited online density control. MonoGS and HI-SLAM2 attain competitive metrics with a small number of Gaussians, but at the cost of the slowest running speed and a long post-optimization stage. Ours achieves comparable quality with MODP, while retaining $\sim$20\% fewer Gaussians, which follows from the Hessian-based online pruning. 

\begin{table}[!t]
\centering
\caption{TUM Real World Dataset Experiment Results}
\label{tab:tum}
\resizebox{0.9\textwidth}{!}{
\begin{tabular}{c|c|c|c|c|c|c|c|c}
\toprule
\hline
& \textbf{RTG-SLAM} & \textbf{GS-ICP} & \textbf{MonoGS} & \textbf{Photo-SLAM} & \textbf{HI-SLAM2} & \textbf{tttLRM} & \textbf{MODP} & \textbf{Ours}  \\ \hline
\textbf{PSNR} $\boldsymbol\uparrow$ & 16.794 & 20.600 & 24.247 & 17.187 & 21.811 & 17.954 & \underline{25.267} & \textbf{25.284} \\ \hline
\textbf{SSIM} $\boldsymbol\uparrow$ & 0.617 & 0.768 & 0.803 & 0.652 & 0.784 & 0.714 & \textbf{0.842} & \underline{0.840} \\ \hline
\textbf{LPIPS} $\boldsymbol\downarrow$ & 0.461 & 0.220 & 0.227 & 0.334 & 0.200 & 0.363 & \textbf{0.187} & \underline{0.194} \\ \hline
\textbf{Running Speed (fps)} & 7.540 & \underline{30.143} & 2.420 & \textbf{46.585} & 4.827 & 12.239 & 11.335 & 11.044 \\ \hline
\textbf{\# of Gaussians} & 303,425 & 1,584,402 & 37,811 & 66,395 & 93,609 & 3,726,362 & 210,275 & 167,234 \\ 
\hline
\bottomrule
\end{tabular}
}
\vspace{-5mm}
\end{table}

\begin{table}[!t]
\centering
\caption{Custom Real World Dataset Experiment Results}
\label{tab:custom}
\resizebox{0.9\textwidth}{!}{
\begin{tabular}{c|c|c|c|c|c|c|c|c}
\toprule
\hline
& \textbf{RTG-SLAM} & \textbf{GS-ICP} & \textbf{MonoGS} & \textbf{Photo-SLAM} & \textbf{HI-SLAM2} & \textbf{tttLRM} & \textbf{MODP} & \textbf{Ours}  \\ \hline
\textbf{PSNR} $\boldsymbol\uparrow$ & 20.460 & 18.503 & 23.387 & 21.254 & 23.640 & 21.316 & \textbf{26.994} & \underline{26.822} \\ \hline
\textbf{SSIM} $\boldsymbol\uparrow$ & 0.718 & 0.740 & 0.793 & 0.757 & 0.832 & 0.748 & \textbf{0.871} & \underline{0.867} \\ \hline
\textbf{LPIPS} $\boldsymbol\downarrow$ & 0.512 & 0.380 & 0.398 & 0.441 & 0.276 & 0.382 & \textbf{0.202} & \underline{0.213} \\ \hline
\textbf{Running Speed (fps)} & 7.275 & \underline{29.081} & 1.424 & \textbf{53.744} & 4.066 & 13.822  & 9.091 & 9.150 \\ \hline
\textbf{\# of Gaussians} & 405,179 & 3,279,379 & 81,868 & 268,928 & 117,082 & 3,725,979 & 385,557 & 309,111 \\ 
\hline \bottomrule
\end{tabular}
}
\vspace{-3mm}
\end{table}

\subsection{Ablation Studies}
\subsubsection{Comparison with Multi-Sample Rendering}
\label{subsubsec:ablation_blur}
\begin{figure*}[!t]
    \centering
    \includegraphics[width=\textwidth]{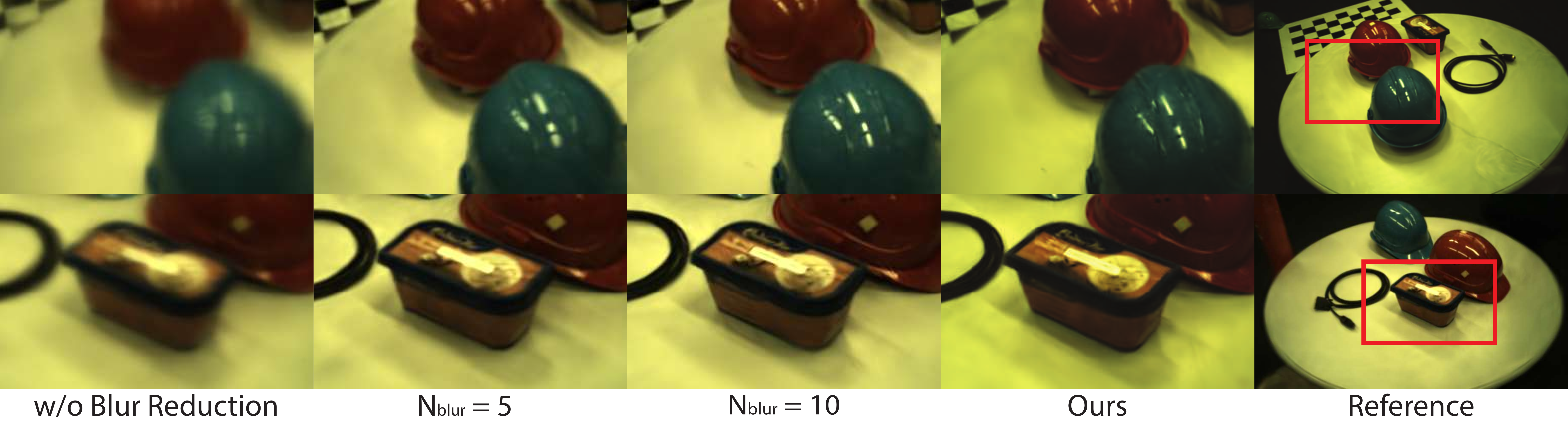}
    \caption{Comparison of exposure-integrated blur with multi-sample blur. $N_\text{blur}=5$ partially reduces motion blur. Compared with $N_\text{blur}=10$, our model provides a competitive result with $2\times$ running speed. }
    \label{fig:ablation_blur}
\vspace{-4mm}
\end{figure*}

To isolate the contribution of our exposure-integrated blur formulation, we compare it against the discrete multi-sample strategy of GSOM~\cite{seiskari2024gaussian} and a no-blur baseline, all within the same framework. We evaluate on the ETH3D dataset with multi-sample variants at $N_\text{blur}=5$ and $N_\text{blur}=10$. \tabref{tab:ablation_blur} reports the quantitative results and \figref{fig:ablation_blur} shows the qualitative comparison.

\paragraph{Rendering Quality.}
The w/o blur reduction baseline yields the worst scores across all metrics, and multi-sampling improves them. Our exposure-integrated blur is slightly behind multi-sampling on \ac{PSNR} and \ac{SSIM} but beats the $N_\text{blur}=5$ variant on \ac{LPIPS}. This perceptual gain arises because the pseudo ground truth is generated by BAD-Gaussians, which itself uses discrete multi-sampling and therefore biases the evaluation. In \figref{fig:ablation_blur}, our model produces cleaner reconstructions than the multi-sample variants while requiring fewer Gaussians to represent the scene; the multi-sample variants spawn additional Gaussians to compensate for the temporal gaps caused by discrete rasterization. 

\paragraph{Running Speed.}
The no-blur baseline is the fastest, while our model incurs an overhead from evaluating the two \ac{ERF} terms of Eq.~\eqref{eq:erf_alpha} during the rasterization pass. Unlike multi-sampling whose cost scales as $\mathcal{O}(N_\text{blur})$, ours runs in $\mathcal{O}(1)$ regardless of $N_\text{blur}$, keeping it viable for online framework.

\subsubsection{Effect of Hessian-based Online Gaussian Pruning}
\begin{figure*}[t!]
    \centering
    \vspace{-4mm}
    \begin{minipage}{0.45\textwidth}
        \centering
        \includegraphics[width=\linewidth]{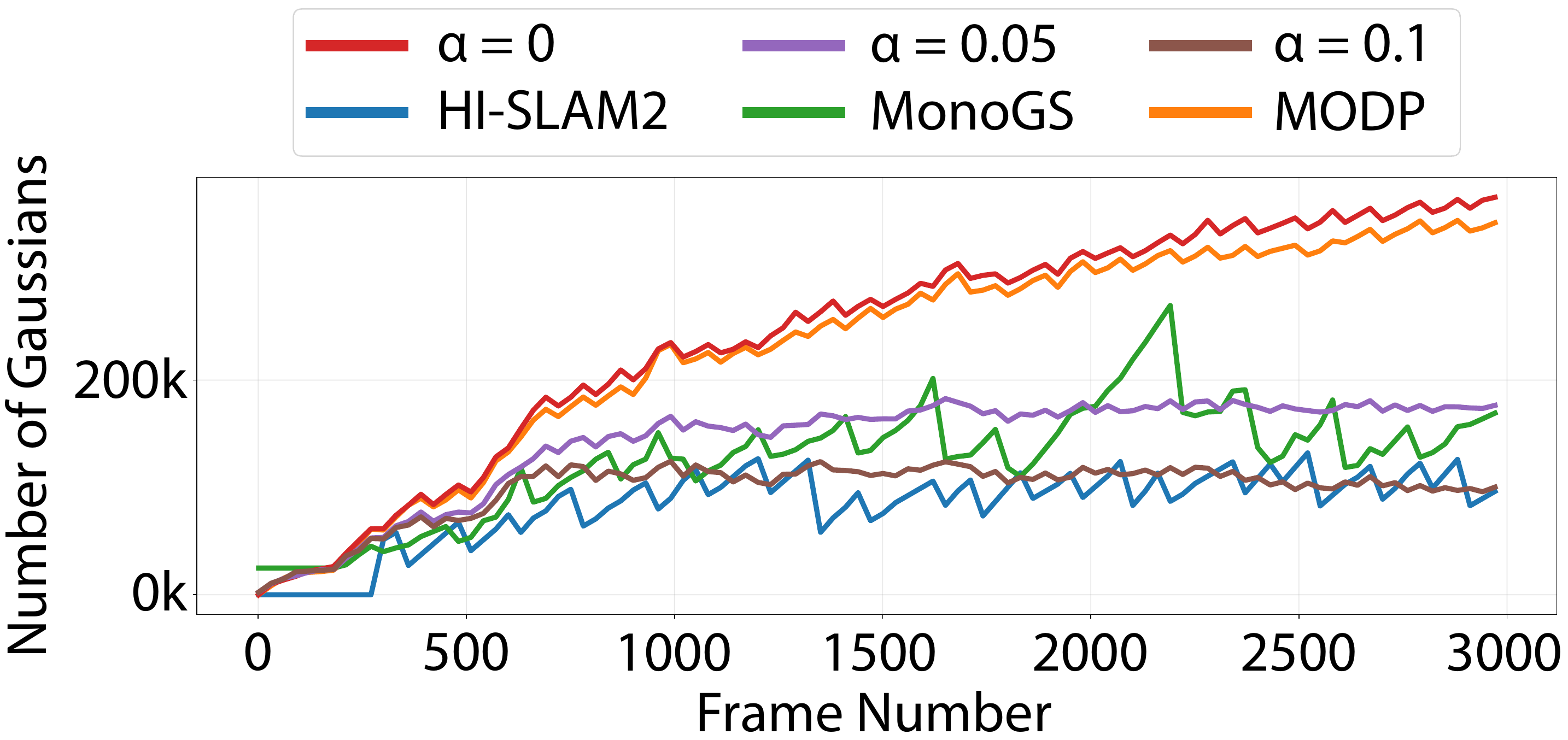}
    \end{minipage}
    \hfill
    \begin{minipage}{0.45\textwidth}
        \centering
        \begin{table}[H]
        \caption{Online Pruning Ratio Comparison}
        \centering
            \resizebox{\linewidth}{!}{
                \begin{tabular}{l|c|c|c|c}
                    \toprule \hline
                     & \textbf{\# of Gaussians} & \textbf{PSNR} \textbf{$\boldsymbol\uparrow$} & \textbf{SSIM} $\boldsymbol\uparrow$ & \textbf{LPIPS} $\boldsymbol\downarrow$ \\ \hline
                    $\alpha = 0 $ & 374,989 & \underline{27.803} & \underline{0.917} & \underline{0.158} \\ \hline
                    $\alpha = 0.05 $ & 170,913 & 27.384 & 0.912 & 0.170 \\ \hline
                    $\alpha = 0.1 $ & \textbf{97,962} & 26.820 & 0.903 & 0.185 \\ \hline \hline
                     MODP & 349,984 & \textbf{27.891} & \textbf{0.919} & \textbf{0.154} \\ \hline
                     HI-SLAM2 & \underline{102,411} & 24.365& 0.886 & 0.189 \\ \hline
                     MonoGS & 148,612 & 24.695 & 0.867 & 0.221 \\ 
                    \hline \bottomrule
                \end{tabular}
            }
            \label{tab:ablation_pruning}
            \vspace{8mm}
        \end{table}
    \end{minipage}
    \vspace{-7mm}
    \caption{Gaussian number growth on dog\_park sequence. The number of Gaussians converges with Hessian-based Online Gaussian Pruning, while the Gaussian number of others grows unbounded or fluctuates unstably. }
    \label{fig:ablation_pruning}
\vspace{-5mm}
\end{figure*}

We ablate the Hessian-based pruning ratio $\alpha \in \{0,~0.05,~0.1\}$ on the dog\_park custom sequence (3000 frames, 4 loops). \figref{fig:ablation_pruning} plots the Gaussian number over frame index and \tabref{tab:ablation_pruning} reports the final counts. For $\alpha=0$ and MODP, the Gaussian number grows without bound, while at $\alpha=0.05$, the count plateaus near frame 1000, around the first loop closure. This confirms that our score retains the Gaussians anchored to each keyframe and drops the redundant ones. For HI-SLAM2 and MonoGS, the Gaussian number fluctuates due to the absence of robust online density control. 

We also run an experiment with a controlled Gaussian count of $\sim$150K for RoMo-GS, compared to Deblurring 3DGS ($\sim$169K). As shown in \tabref{tab:Deblur_NeRF}, increasing the Gaussian count consistently improves the quantitative metrics, while LPIPS remains higher than Deblurring 3DGS. We attribute this gap to per-Gaussian time averaging before alpha compositing, which may slightly smooth object boundaries and thus penalize LPIPS. 

\begin{table}[!t]
\centering
\caption{Comparison with Multi-Sample Rendering on ETH3D Dataset}
\label{tab:ablation_blur}
\resizebox{0.9\columnwidth}{!}{
\begin{tabular}{c|c|c|c|c}
\toprule
\hline
& \textbf{Ours (Exposure-Integrated)} & \textbf{w/o Blur Reduction} & \textbf{Multi Sample} ($N_\text{blur}=5$) & \textbf{Multi Sample} ($N_\text{blur}=10$)\\ \hline
\textbf{PSNR} $\boldsymbol\uparrow$ & 25.367 & 24.355 & \underline{25.951} & \textbf{26.003} \\ \hline
\textbf{SSIM} $\boldsymbol\uparrow$ & \underline{0.905} & 0.892 & \textbf{0.911} & \textbf{0.911} \\ \hline
\textbf{LPIPS} $\boldsymbol\downarrow$ & \underline{0.113} & 0.202 & 0.121 & \textbf{0.107} \\ \hline
\textbf{Running Speed (fps)} & \underline{3.695} & \textbf{5.698} & 2.448 & 1.592 \\ \hline
\textbf{\# of Gaussians} & \textbf{117,669} & 129,128 & \underline{124,831} & 126,902 \\ 
\hline
\bottomrule
\end{tabular}
}
\vspace{-3mm}
\end{table}

\vspace{-4mm}
\begin{table}[hbt!]
\centering
\caption{Experiment with $\sim$150K Gaussians on the Deblur-NeRF Synthetic Dataset}
\label{tab:Deblur_NeRF}
\resizebox{0.6\columnwidth}{!}{%
\begin{tabular}{l|c|c|c|c} \toprule
& \textbf{PSNR} & \textbf{SSIM} & \textbf{LPIPS} & \textbf{\# of Gaussians} \\
\midrule
\textbf{Ours} & \underline{25.148} & \underline{0.799} & 0.179 & 75,813 \\
\textbf{Ours (150K Gaussians)} & \textbf{25.499} & \textbf{0.815} & \underline{0.164} & 150,164 \\
\textbf{Deblurring 3DGS} & 24.174 & 0.768 & \textbf{0.097} & 168,837
\\ \bottomrule
\end{tabular}%
}
\vspace{-4mm}
\end{table}

\subsubsection{Additional Comparison with MODP on Blurry Scene}
\begin{figure}[t!]
    \centering
    \includegraphics[width=0.8\columnwidth]{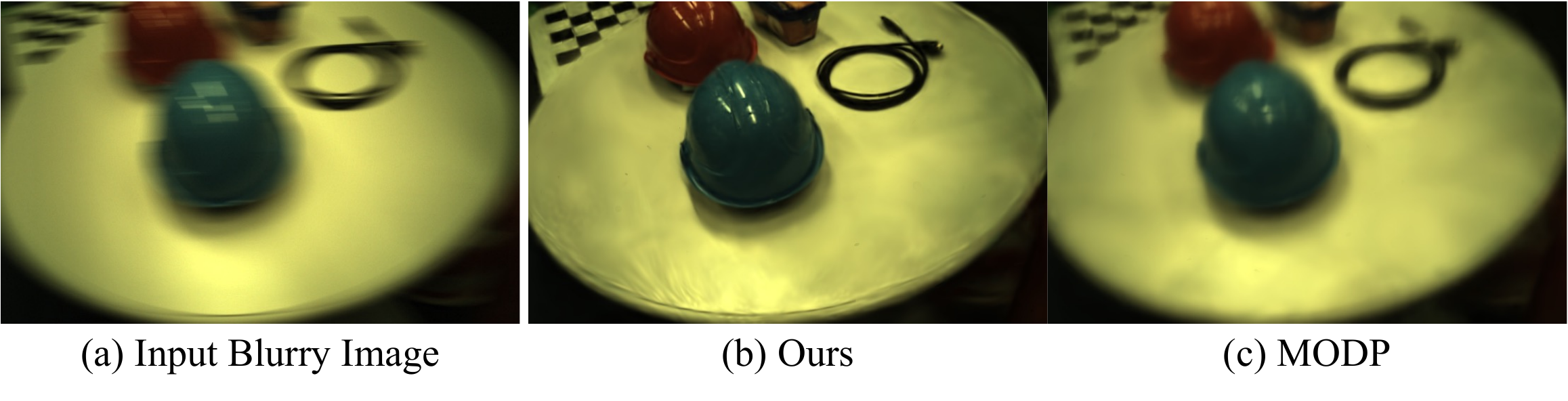}
    \vspace{-2mm}
    \caption{RoMo-GS and MODP on the blurry input (camera\_shake\_1 seq of ETH3D). }
    \label{fig:MODP_exp}
    \vspace{-5mm}
\end{figure}

On \tabref{tab:tum} and \tabref{tab:custom}, RoMo-GS matches MODP while using $\sim$20 fewer Gaussians. To show clear qualitative improvements of RoMo-GS on severe motion blur, we report an additional experiment on the ETH3D dataset. As shown by the comparison at \tabref{tab:MODP_exp} and \figref{fig:MODP_exp}, RoMo-GS is particularly beneficial for online mapping under rapid camera motion. 

\begin{table}[t!]
\centering
\caption{Comparison with MODP on the camera\_shake\_1 sequence of ETH3D}
\label{tab:MODP_exp}
\resizebox{0.5\columnwidth}{!}{%
\begin{tabular}{l|c|c|c|c} \toprule
& \textbf{PSNR} & \textbf{SSIM} & \textbf{LPIPS} & \textbf{\# of Gaussians} \\
\midrule
\textbf{Ours} & \textbf{25.648} & \textbf{0.908} & \textbf{0.113} & \textbf{106,285} \\
\textbf{MODP} & 23.726 & 0.879 & 0.244 & 129,165
\\ \bottomrule
\end{tabular}%
}
\vspace{-3mm}
\end{table}

\subsubsection{Complete System Ablation Study}
We additionally ablate deblurring, pruning, joint pose/velocity, and the online \ac{LM} step to evaluate the complete RoMo-GS pipeline on the ETH3D dataset as in \tabref{tab:Add_Ablation}. The evaluation metrics get worse when we remove each module of RoMo-GS, except for Hessian-based pruning. 

\vspace{-4mm}
\begin{table}[hbt!]
\centering
\caption{Complete system ablation study of RoMo-GS on the camera\_shake\_1 sequence of ETH3D}
\label{tab:Add_Ablation}
\resizebox{0.8\columnwidth}{!}{%
\begin{tabular}{c|c|c|c||c|c|c|c} \toprule
Pruning & Deblurring & LM & Pose/velocity & \textbf{PSNR} & \textbf{SSIM} & \textbf{LPIPS} & \textbf{\# of Gaussians} \\
\midrule
\cmark & \cmark & \cmark & \cmark & 24.145 & 0.876 & 0.232 & 62,582 \\
\xmark & \cmark & \cmark & \cmark & \textbf{25.648} & \textbf{0.908} & \textbf{0.113} & 106,285 \\
\xmark & \xmark & \cmark & \cmark & 23.485 & 0.876 & 0.247 & 139,051 \\
\xmark & \xmark & \xmark & \cmark & 23.457 & 0.876 & 0.250 & 131,815 \\
\xmark & \xmark & \xmark & \xmark & 23.059 & 0.872 & 0.251 & 122,100 \\ \bottomrule
\end{tabular}%
}
\end{table}
\vspace{-4mm}

\subsubsection{Tracking Accuracy}
We additionally evaluate the tracking accuracy on the TUM~\cite{sturm12iros} dataset. Specifically, we compare the \ac{RMSE} of the \ac{ATE} from the Gaussian-jointly-optimized poses against native ORB-SLAM3~\cite{campos2021orb}, which serves as the input pose stream to our mapper. On average, \ac{ATE}$_t$ is reduced by about \textbf{2\%} and \ac{ATE}$_r$ by about \textbf{0.4\%}. While the improvement is modest, it confirms that jointly optimizing the camera state with the Gaussian map not only sharpens rendering but also provides a marginal refinement of the front-end pose estimates. 
\section{Conclusion}
\label{sec:conclusion}
We present \textbf{RoMo-GS}, an efficient online \ac{3DGS} system designed for robotic mapping. Replacing the $N$-sample averaging of prior offline global motion-blur-aware \ac{3DGS} methods, we analytically integrate each projected Gaussian's contribution over the exposure window into a single \ac{ERF} evaluation, running with $\mathcal{O}(1)$ complexity per Gaussian. Around this kernel, we jointly refine pose, velocity, and Gaussians with an Adam loop followed by an online \ac{LM} step, and bound the map size by online density control driven by a Hessian-based score. On the Deblur-NeRF synthetic dataset and the ETH3D real-world dataset, RoMo-GS matches or exceeds offline comparisons while running 15--150$\times$ faster and remaining online; on TUM and our self-collected multi-platform dataset, RoMo-GS remains competitive with $\sim20$\% fewer Gaussians than the online baseline MODP~\cite{wu2025monocular}. 
\section{Limitation}
\label{sec:limitation}
Our derivation assumes a linear motion model, which cannot capture curved intra-exposure trajectories, so extending with $\mathfrak{se}(3)$ Lie algebra is required. The proposed motion blur model also requires the exposure time $t_e$ as a prior; jointly optimizing $t_e$ or reading it from camera metadata would address this. Finally, the front-end tracker constrains mapping quality, which can lead to back-end reconstruction failure when the front-end vision tracker fails. Tighter coupling between the \ac{3DGS} map and the tracker may provide more accurate results and prevent front-end tracker failure. 
% \newpage
\newpage

\section*{ACKNOWLEDGMENT}
The authors thank and acknowledge ETH Zürich for allowing us to use their experimental space for real-world testing, data collection, and demo video filming. 

\balance
\bibliography{references}

\newpage
\appendix
\section*{Appendices}
\appendix
\section{Derivation of the Exposure-Integrated Motion Blur}
\label{app:blur}

Starting from Eq.~\eqref{eq:erf_integral}, the quadratic exponent can be written as $a\tau^2+2b\tau+c$, where $a\ge0$ measures how fast the Mahalanobis exponent grows along the Gaussian's sweep direction and $c$ is the standard \ac{3DGS} exponent of the stationary Gaussian. For $a>0$, let $\tau^*=-b/a$ denote the closest-approach time. The exponent then attains its minimum $c-b^2/a$, and completing the square gives
\begin{equation}
    a\tau^2+2b\tau+c=a(\tau-\tau^*)^2+(c-\frac{b^2}{a}),
    \label{eq:complete_square}
\end{equation}
which factors out the constant term from Eq.~\eqref{eq:erf_integral} and leaves a shifted Gaussian integral over the exposure window. Substituting $u=\sqrt{a/2}(\tau-\tau^*)$ and using $\operatorname{erf}(u) = 2/\sqrt{\pi} \int_0^u \exp (-t^2)dt$,
\begin{equation}
    \int_{-t_e/2}^{t_e/2}\exp\bigl(-\frac{a}{2}(\tau-\tau^*)^2\bigr)d\tau
    =\sqrt{\frac{\pi}{2a}}\,[\operatorname{erf}(u_+)-\operatorname{erf}(u_-)],
    \label{eq:erf_substitution}
\end{equation}
with $u_\pm=\sqrt{a/2}(\pm t_e/2-\tau^*)$. Combining the prefactor $o_j/t_e$ of Eq.~\eqref{eq:erf_integral} with the constant $\exp(-(c-b^2/a)/2)$ from Eq.~\eqref{eq:complete_square} and the \ac{ERF} difference from Eq.~\eqref{eq:erf_substitution} yields Eq.~\eqref{eq:erf_alpha}.

\paragraph{Geometric Interpretation.}

Eq.~\eqref{eq:erf_alpha} factors into three terms, $P=(o_j/t_e)\sqrt{\pi/(2a)}$, $G=\exp(-(c-b^2/a)/2)$, and $E=\operatorname{erf}(u_+)-\operatorname{erf}(u_-)$, each of which admits an intuitive reading. The prefactor $P$ normalizes the sweep by its effective duration: $\sqrt{1/a}$ scales inversely with how fast the Gaussian crosses the pixel, so $P$ grows when the Gaussian lingers near $\mathbf{x}$ and shrinks when it sweeps past quickly. The exponential $G$ is the kernel value at the Mahalanobis closest-approach time $\tau^*$ along the projected trajectory. The \ac{ERF} difference $E$ measures the normalized Gaussian mass captured within the finite exposure window; it approaches $2$ when the window covers nearly all mass around $\tau^*$ and shrinks toward zero when this mass lies far outside $[-t_e/2,t_e/2]$. Thus, $P\,G\,E$ decomposes the analytic footprint into a duration-normalized amplitude, a peak kernel value, and an exposure-window coverage term.

\section{Numerical Stability near the Stationary Limit}
\label{app:stability}

The exposure-integrated footprint of Eq.~\eqref{eq:erf_alpha} contains the prefactor $P=(o_j/t_e)\sqrt{\pi/(2a)}$, which diverges as the Mahalanobis sweep coefficient $ a = {\mathbf v'_{i,j}}^\top {\boldsymbol\Sigma'_{i,j}}^{-1}\,\mathbf v'_{i,j} $ approaches zero, even though the integrand of Eq.~\eqref{eq:erf_integral} remains finite. Physically, $a\to 0$ occurs when the projected pixel velocity $\mathbf v'_{i,j}$ vanishes or when the Gaussian is highly elongated along the sweep direction, making the Mahalanobis growth along that direction nearly zero. In these cases, the Gaussian is effectively stationary on the image plane over the exposure window.

\paragraph{Continuity at $\mathbf a \boldsymbol \to \mathbf 0 $.}
The apparent singularity of $P$ is removable. Recall that the quadratic exponent in Eq.~\eqref{eq:erf_integral} can be written as $a\tau^2 + 2b\tau + c $. 
For $a>0$, the error-function arguments in Eq.~\eqref{eq:erf_alpha} are
\begin{equation}
    u_\pm
    =
    \sqrt{\frac{a}{2}}
    \left(\pm\frac{t_e}{2}-\tau^*\right),
    \qquad
    \tau^*=-\frac{b}{a}.
\end{equation}
Equivalently, we can write
\begin{equation}
    u_\pm = u_0 \pm \Delta,
    \qquad
    u_0 = \frac{b}{\sqrt{2a}},
    \qquad
    \Delta = \frac{t_e}{2}\sqrt{\frac{a}{2}} .
\end{equation}
For $a>0$, since ${\boldsymbol\Sigma'}^{-1}$ is positive definite, Cauchy--Schwarz gives $b^2 \le ac$. Therefore $u_0^2=b^2/(2a)\le c/2$, so $u_0$ remains bounded as $a\to0$.
\begin{align}
    E
    &=
    \operatorname{erf}(u_+) - \operatorname{erf}(u_-)
    =
    \operatorname{erf}(u_0+\Delta)-\operatorname{erf}(u_0-\Delta) \nonumber \\
    &=
    \frac{4}{\sqrt{\pi}} e^{-u_0^2}\Delta + \mathcal{O}(\Delta^3)
    =
    \frac{2}{\sqrt{\pi}}\,t_e\sqrt{\frac{a}{2}}\,
    \exp\!\left(-\frac{b^2}{2a}\right)
    + \mathcal{O}(a^{3/2}).
\end{align}
Substituting this expression into Eq.~\eqref{eq:erf_alpha}, the $\sqrt{a}$ factors cancel and the exponential terms combine as
\begin{equation}
    \exp\!\left(-\frac{c-b^2/a}{2}\right)
    \exp\!\left(-\frac{b^2}{2a}\right)
    =
    \exp\!\left(-\frac{c}{2}\right).
\end{equation}
Hence,
\begin{equation}
    \lim_{a\to 0}\bar\alpha_{i,j}(\mathbf x)
    =
    o_j\,\exp\!\Bigl(
        -\frac{1}{2}\,
        \boldsymbol\delta^\top{\boldsymbol\Sigma'_{i,j}}^{-1}\boldsymbol\delta
    \Bigr)
    =
    \alpha_{i,j}(\mathbf x,\boldsymbol\mu'_{i,j}),
    \label{eq:erf_limit}
\end{equation}
recovering the standard static \ac{3DGS} footprint. This is consistent with the original integral: when $a=0$, the projected sweep has zero Mahalanobis length, the quadratic exponent becomes constant over the exposure interval, and Eq.~\eqref{eq:erf_integral} collapses to $o_j\exp(-c/2)$.

\paragraph{Practical Implementation.}
Although Eq.~\eqref{eq:erf_alpha} is continuous at the limit, evaluating it in single precision at small $a$ is unstable: $\sqrt{1/a}$ amplifies floating-point noise, $b/a$ produces $0/0$-type cancellations, and $\operatorname{erf}(u_\pm)$ loses significant digits when $|u_\pm|\ll 1$. We therefore branch in the rasterizer on a tolerance $\varepsilon_a$:
\begin{equation}
    \bar\alpha_{i,j}(\mathbf x)
    \;=\;
    \begin{cases}
        P\,G\,E & \text{if } a > \varepsilon_a, \\
        o_j\,\exp\bigl(-\frac{1}{2}\,\boldsymbol\delta^\top {\boldsymbol\Sigma'_{i,j}}^{-1}\boldsymbol\delta\bigr) & \text{otherwise},
    \end{cases}
    \label{eq:branch}
\end{equation}
with $\varepsilon_a = 10^{-6}\,\mathrm{px}^2$ in our implementation. The static branch is mathematically the $a\to 0$ limit of the analytically integrated motion-blur model expression by Eq.~\eqref{eq:erf_limit}, so the rendered image is continuous across the branch boundary. In addition, we (i) reject samples with numerically negative quadratic-form values ($c<0$) caused by finite-precision conic round-off, (ii) skip Gaussians whose minimum exponent $(c-b^2/a)/2$ exceeds a cut-off so that $G$ underflows to zero, and (iii) clamp $\bar\alpha_{i,j}\le 1-\varepsilon$ before front-to-back compositing, matching the safeguards of standard \ac{3DGS}~\cite{kerbl3Dgaussians}.

\paragraph{Gradient Continuity.}
The same branch is applied in the backward pass to keep the gradients well-behaved near the stationary limit. On the static branch, $\partial\bar\alpha/\partial\boldsymbol\mu',~\partial\bar\alpha/\partial{\boldsymbol\Sigma'}^{-1}$ reduce to the standard \ac{3DGS} VJPs, while $\partial\bar\alpha/\partial\mathbf v'_{i,j}=\mathbf 0$ on this branch because the rendered opacity is evaluated by the static footprint. Empirically, this guarantees that Gaussians whose projected velocity transiently passes through zero, common for far-field Gaussians or for keyframes captured during near-stationary segments of the trajectory, still receive valid gradients from the geometric residual and continue to optimize as ordinary static primitives.

\section{LM Refinement Details}
\label{app:lm}
Let $\boldsymbol\theta$ collect the Gaussian parameters and $\mathbf r$ denote the per-pixel residual under the standard \ac{3DGS} loss, with Huber weights collected in $\mathbf W$. The damped Gauss-Newton normal equation is
\begin{equation}
  \bigl(\mathbf J^\top\mathbf W\mathbf J
   +\lambda\,\mathrm{diag}(\mathbf J^\top\mathbf W\mathbf J)\bigr)\,
  \delta\boldsymbol\theta
  \;=\; -\,\mathbf J^\top\mathbf W\mathbf r,    
\end{equation}
solved by preconditioned conjugate gradient with the diagonal of $\mathbf J^\top\mathbf W\mathbf J$ as the preconditioner. $\lambda$ is initialized per-keyframe as $\lambda_0=\tau\cdot\max_k[\mathrm{diag}(\mathbf J^\top\mathbf W\mathbf J)]_k$ with $\tau=10^{-3}$, following Nielsen~\cite{nielsen1999damping}, and updated by the standard accept/reject rule.

\section{Online Hessian-Based Pruning Details}
\label{app:pruning}
\paragraph{Chain rule.}
Under standard \ac{3DGS} compositing, the opacity-axis gradient for Gaussian $j$ at pixel $\mathbf x$ in keyframe $i$ factors as
\begin{equation}
  \frac{\partial L_i}{\partial o_j} = g_{i,j}(\mathbf{x}) \frac{\partial L}{\partial \alpha_{i,j}(\mathbf{x} )},
  \qquad
  g_{i,j}(\mathbf x)=\exp \Bigl(-\frac{1}{2} (\boldsymbol{\mu}'_{i,j}-\mathbf{x})^\top {\boldsymbol{\Sigma}'_{i,j}}^{-1} (\boldsymbol{\mu}'_{i,j}-\mathbf{x})\Bigr),
\end{equation}
where $g_{i,j}$ is the screen-space Gaussian kernel value and $\partial L/\partial\alpha$ carries the standard front-to-back compositing chain. Summing the squares over visible pixels and active keyframes gives the score in the main text.

\paragraph{Comparison with Speedy-Splat.}
Speedy-Splat scores Gaussians by the squared kernel-value sensitivity of the rendered image, $\sum_p(\partial I/\partial G_{i,p})^2$, extracted by an offline backward pass over all training views. Our score is the opacity-axis Gauss-Newton diagonal of the same training loss and is read off the gsplat backward, already executed during online optimization, adding no rendering or extra
backward cost.

\begin{figure*}[h]
    \centering
    \includegraphics[width=\textwidth]{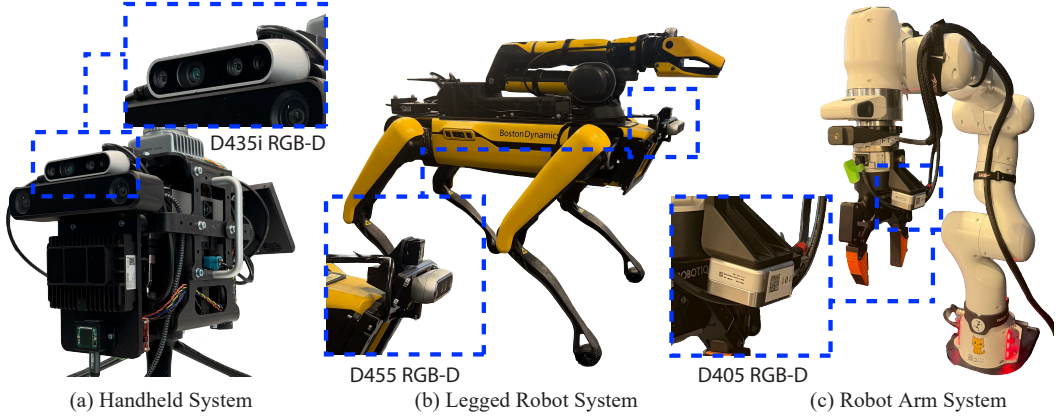}
    \caption{Real-world robot platforms used to acquire the custom dataset. (a) Handheld system with an Intel RealSense D435i RGB-D camera. (b) Spot quadruped robot with an Intel RealSense D455 RGB-D camera. (c) Franka robot arm with an Intel RealSense D405 RGB-D camera.}
    \vspace{-5mm}
    \label{fig:custom_system}
\end{figure*}

\section{Custom Dataset}
\label{app:dataset}
For a broader evaluation, we acquire a custom dataset using three real-world robot platforms: a handheld rig, a legged robot, and a robot arm. The handheld and legged-robot systems cover large-scale scenes under different motion patterns and viewing directions, while the robot-arm system captures classical tabletop scenes with controlled manipulator motion. \figref{fig:custom_system} shows the hardware setup for each platform. The dataset contains seven sequences in total: two handheld sequences, two legged-robot sequences, and three robot-arm sequences.
\begin{itemize}
    \item \textbf{Stair:} Captured in a large stair area using the handheld system. The sequence includes motion blur induced by human locomotion and vertical camera motion.

    \item \textbf{Kitchen:} Captured in a kitchen area using the handheld system. The sequence includes motion blur induced by human locomotion and horizontal camera motion.

    \item \textbf{Dog\_Park:} Captured in a legged-robot experimental area using a Spot quadruped robot. The robot traverses the same area in four loops, producing legged-locomotion induced motion blur.

    \item \textbf{Corridor:} Captured in a square corridor containing office and kitchen areas using a Spot quadruped robot. The robot completes one loop, and the sequence includes motion blur induced by legged locomotion.

    \item \textbf{Arm1, Arm2, Arm3:} Captured on classic tabletop scenes with various objects using a Franka robot arm. These sequences include robot-arm-induced motion blur.
\end{itemize}

\end{document}